\documentclass[runningheads]{llncs}

 \usepackage{eccv}

\usepackage{orcidlink}
\usepackage{pifont}
\usepackage{multirow}
\usepackage[table]{xcolor}
\usepackage[T1]{fontenc}
\usepackage{lmodern}
\usepackage{microtype}
\usepackage{cmap}

\usepackage{eccvabbrv}

\usepackage{graphicx}
\usepackage{booktabs}

\usepackage[accsupp]{axessibility}
\usepackage{pifont}

\usepackage{hyperref}

\begin{document}

\title{RePer-360: Releasing Perspective Priors for 360$^\circ$ Depth Estimation via Self-Modulation} 

\titlerunning{RePer-360}

\author{Cheng~Guan\inst{1,2} \and
 Chunyu~Lin\inst{1,2} \and
Zhijie~Shen\inst{1,2}\textsuperscript{\ding{41}} \and
Junsong~Zhang\inst{1,2} \and
Jiyuan~Wang\inst{1,2}
}

\authorrunning{Guan et al.}

\institute{\small Institute of Information
Science, Beijing Jiaotong University, Beijing 100044, China. \and
Visual Intelligence +X International Cooperation Joint Laboratory of MOE,
Beijing 100044, China.
\ding{41} Corresponding author: Zhijie Shen
\email{\{24120278,zjshen\}@bjtu.edu.cn}}

\maketitle

\begin{abstract}
Recent depth foundation models trained on perspective imagery achieve strong performance, yet generalize poorly to 360$^\circ$ images due to the substantial geometric discrepancy between perspective and panoramic domains. Moreover, fully fine-tuning these models typically requires large amounts of panoramic data. To address this issue, we propose RePer-360, a distortion-aware self-modulation framework for monocular panoramic depth estimation that adapts depth foundation models while preserving powerful pretrained perspective priors. Specifically, we design a lightweight geometry-aligned guidance module to derive a modulation signal from two complementary projections (i.e., ERP and CP) and use it to guide the model toward the panoramic domain without overwriting its pretrained perspective knowledge. We further introduce a Self-Conditioned AdaLN-Zero mechanism that produces pixel-wise scaling factors to reduce the feature distribution gap between the perspective and panoramic domains. In addition, a cubemap-domain consistency loss further improves training stability and cross-projection alignment. By shifting the focus from complementary-projection fusion to panoramic domain adaptation under preserved pretrained perspective priors, RePer-360 surpasses standard fine-tuning methods while using only 1\% of the training data. Under the same in-domain training setting, it further achieves an approximately 20\% improvement in RMSE. The code is available at \url{https://github.com/munimo/RePer360}.

\end{abstract}

\section{Introduction}
\begin{figure}[t]
  \centering  \includegraphics[width=0.55\textwidth]{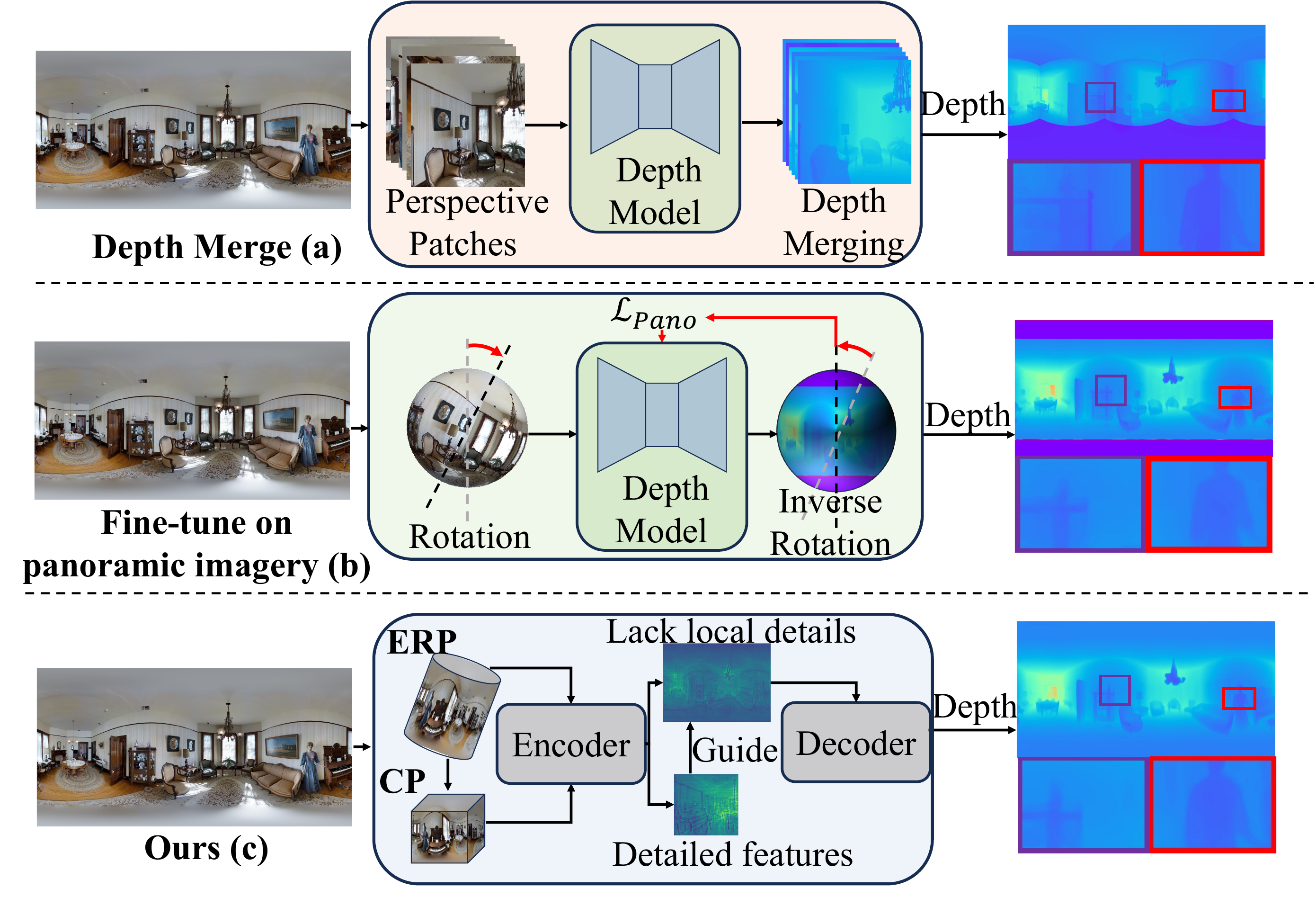} 
    \caption{Technical strategies based on pretrained perspective models (Persp): (Top) Patch fusion suffers from artifacts; (Middle) End-to-end fine-tuning requires massive data; (Bottom) Our RePer-360 enables precise detail transfer with minimal data.}
\label{fig:Introduction}
\end{figure}
Unlike traditional perspective images, 360$^\circ$ imagery captures a complete spherical environment in a single shot, enabling holistic scene understanding and broad applications in virtual reality and autonomous systems.
Recent depth foundation models, such as Depth Anything Models~\cite{DAMv1,DAMv2} (DAMs), generalize well on perspective images but degrade markedly on $360^\circ$ panoramas. We attribute this drop to a severe prior mismatch: pretrained representations follow perspective-domain statistics that panoramic distortions violate.

To address this challenge, existing solutions can be broadly categorized into two lines of work (see Fig.~\ref{fig:Introduction}). The first follows a \emph{projection-based inference-and-fusion} paradigm. MoGe-2~\cite{wang2025moge} adopts this strategy by partitioning a panorama into multiple perspective views, performing inference per view, and fusing the predictions back to the panoramic space. Building upon this pipeline, ST$^2$360D~\cite{cao2025st} organizes perspective projections into a pseudo-temporal sequence and leverages video foundation models to further improve cross-view consistency. While such pipelines can alleviate local artifacts, they typically treat projected views as approximately independent and do not explicitly model global spherical geometry. Moreover, multi-view inference and the subsequent fusion steps introduce additional computational overhead and inference latency, limiting efficiency.

The second line fine-tunes perspective depth models in the panoramic domain using additional $360^\circ$ data. For instance, PanDA~\cite{cao2025panda} employs LoRA for parameter-efficient adaptation under a semi-supervised framework. Although this approach can better preserve global geometric consistency, it often relies on large-scale panoramic data. More importantly, without explicitly modeling panoramic distortions, domain adaptation under limited supervision may fail to effectively transfer pretrained perspective priors and can even overwrite them or induce representation drift, ultimately harming generalization.

Inspired by multi-projection strategies, we first explored a complementary-projection feature fusion scheme: we froze the backbone of a depth foundation model, extracted two sets of features from complementary projections, such as Cubemap Projection (CP) and ERP, and applied established fusion strategies to simultaneously handle distortion and transfer perspective priors. However, we observed only marginal gains. We hypothesize that explicit fusion perturbs the pretrained feature statistics, which in turn degrades performance (see Sec.~\ref{sec:ablation}). Motivated by this finding, we argue that complementary projection features should \emph{not} be hard-fused into a new representation; instead, the complementary features should be used as a structured guidance signal to enable more stable domain transfer of pretrained priors.

To this end, we propose \textbf{RePer-360}, a distortion-aware \emph{self-modulation} framework for panoramic monocular depth estimation. Rather than performing direct feature fusion, RePer-360 aligns cross-domain feature distributions via \emph{normalization-based modulation}, thereby adapting to panoramic distortions while avoiding overwriting or damaging pretrained perspective priors. Concretely, we propose \emph{Self-Conditioned AdaLN-Zero} (SCAdaLN-Zero), a zero-initialized normalization modulator that derives pixel-wise scaling from geometry-aligned guidance and applies it within normalization layers to enable stable adaptation. A lightweight geometry-aligned guidance module extracts modulation cues from complementary projections (CP and ERP), explicitly encoding distortion differences and geometric correspondences. In addition, we design a \emph{cubemap-domain consistency loss} to mitigate distortion-induced imbalance and to enhance cross-projection consistency.

RePer-360 achieves high data efficiency and superior accuracy: using only $\sim$1\% of the training data of prior methods (1k vs.\ 120k image pairs), it surpasses the previous state of the art; under the same training data scale, it improves RMSE by up to 22.4\% relatively. Our contributions are summarized as follows:
\begin{itemize}
\item We reformulate panoramic adaptation as \emph{distortion-aware, guidance-based domain adaptation}, using complementary projections as guidance for prior-preserving transfer instead of hard fusion.
\item We propose RePer-360, a distortion-aware self-modulation framework that performs stable perspective-to-panorama alignment via geometry-aligned guidance and normalization-based modulation.
\item Extensive experiments demonstrate that our method is efficient under limited panoramic supervision and yields consistent performance improvements.
\end{itemize}

\section{Related work}
\label{gen_inst}
\subsection{Monocular 360 Depth Estimation}
Although a giant leap has been made in the 3D area~\cite{jiyuan2,jiyuan3,jiyuan4,jiyuan6,jiyuan5}, panoramic distortions significantly degrade the performance of depth estimation models originally designed for perspective images. Early research addressed this challenge through two main approaches: developing fusion frameworks~\cite{Yoon2021SphereSRI,Ai_2024_CVPR,shen2024revisiting} that integrate Cubemap projection (CP) with Equirectangular projection (ERP), and designing specialized attention modules~\cite{Shen2022PanoFormerPT,Yun2023EGformerEG,zhang2025sgformer} to capture distortion patterns directly in the spherical domain.
With the development of large pretrained perspective models such as Depth Anything (DAM), research has shifted toward adapting these powerful priors to panoramic scenarios. Current methodologies follow two technical pathways. 
The first category divides panoramic images into perspective-aligned patches to alleviate distortions. This strategy is exemplified by MoGe-2~\cite{wang2025moge} and further extended by ST$^2$360D~\cite{cao2025st}, which arranges projected views into pseudo-temporal sequences and leverages video-based foundation models for improved consistency. Similar paradigms are adopted by OmniFusion~\cite{Li2022OmniFusion3M} and HRDFuse~\cite{Ai2023HRDFuseM3}. Although these methods mitigate local inconsistencies, they process projected views independently without explicitly modeling global spherical geometry. Moreover, their reliance on stitching procedures or multi-frame inference introduces substantial computational overhead and increased latency.
The second category focuses on direct fine-tuning of perspective foundation models using large-scale panoramic datasets. PanDA~\cite{cao2025panda} employs LoRA~\cite{hu2022lora} for parameter-efficient semi-supervised fine-tuning of DAM, incorporating a Möbius transformation-based spatial augmentation strategy. Depth Anywhere~\cite{wang2024depth} instead adopts an indirect distillation scheme, generating pseudo-labels from a perspective DAM to supervise panoramic models. However, these approaches share a common limitation: the underlying architectures lack an inherent design for panoramic geometry. Consequently, their performance heavily depends on large-scale 360$^\circ$ supervision, which increases data requirements and may weaken high-fidelity detail preservation. Distortion-aware domain adaptation has also been studied in related panoramic perception tasks such as semantic segmentation~\cite{zheng2023look,zheng2023both,zhang2022bending}.

\subsection{Diffusion Transformer}
The Diffusion Transformer~\cite{DiT} (DiT) leverages Adaptive Normalization~\cite{perez2018film} (AdaLN) for feature-level conditioning, a mechanism also widely adopted in GANs~\cite{brock2018large,karras2019style} and U-Net based diffusion models~\cite{dhariwal2021diffusion}. Instead of using fixed parameters, the AdaLN module in DiT dynamically generates channel-level scaling and shifting parameters directly from timestep and class embeddings. A key enhancement introduced in DiT is AdaLN-Zero, which incorporates additional scaling factors that are initialized to zero and applied before residual connections, significantly improving training stability. While these methods condition on externally predefined signals such as timesteps or class embeddings, our approach derives conditioning signals from internally computed, geometry-aligned cross-projection features. This shift replaces generic sequential conditioning with task-specific structural cues tailored to panoramic adaptation.

\section{Methodology}
\label{headings}
\begin{figure*}[t]
  \centering
  \includegraphics[width=0.95\textwidth]{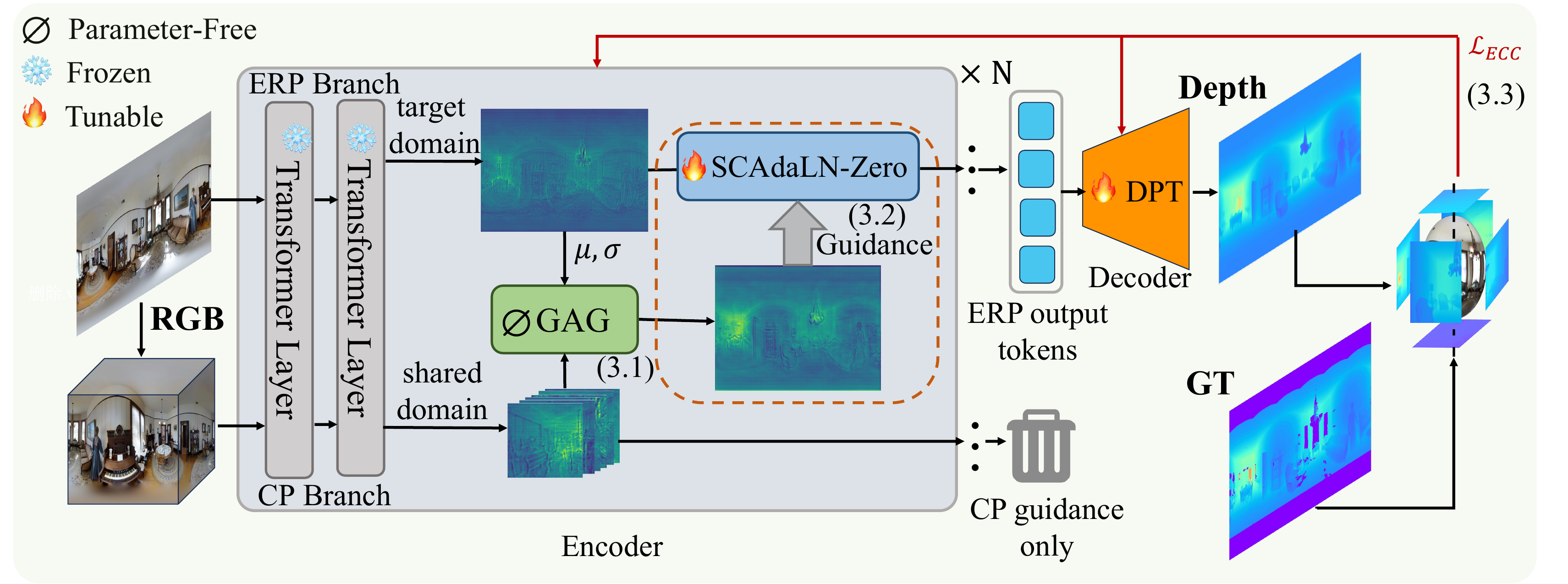} 
  \caption{Overview of the proposed framework. Our framework takes a single panorama as input and outputs the corresponding depth map. The network leverages the rich visual priors in the DAM to make it suitable for panoramic depth estimation.} 
  \label{fig:framework}
\end{figure*}
\noindent \textbf{Overview.} RePer-360 adapts a pretrained depth foundation model to the panoramic domain while preserving its priors. As illustrated in Fig.~\ref{fig:framework}, our framework follows a guidance–modulation–supervision pipeline.
The panorama is first processed by a frozen DAM encoder to extract ERP features, while a cubemap projection (CP) branch provides complementary perspective-aligned features. A Geometry-Aligned Guidance (GAG) module (Sec.~\ref{sec:GAG}) aligns and aggregates ERP and CP representations to produce geometry-aware conditioning signals.
These signals are injected into normalization layers via a Self-Conditioned AdaLN-Zero module (Sec.~\ref{sec:SCAdaLN}), enabling structured distortion-aware adaptation while preserving pretrained representations.
To stabilize learning under spherical distortion, we further introduce an E2C Consistency Loss (ECCLoss) (Sec.~\ref{sec:ECC}), which enforces cross-projection geometric consistency in the cubemap domain.
Through guidance-driven normalization, RePer-360 reframes panoramic depth estimation as structured prior adaptation rather than explicit feature fusion.
\subsection{Geometry-Aligned Guidance for Stable Modulation}
\label{sec:GAG}

We introduce a Geometry-Aligned Guidance (GAG) module to generate auxiliary guidance for SCAdaLN-Zero modulation, rather than replacing the backbone feature. Since the backbone is pretrained on perspective images, each cubemap projection (CP) face is closer to the backbone's training distribution than the distorted ERP image. Therefore, CP features are more geometrically consistent with the backbone representation and provide reliable local structural details. However, CP features are split into six faces and suffer from boundary discontinuities, making them unsuitable as a global representation.
\begin{figure}[t]
\centering
\includegraphics[width=0.55\textwidth]{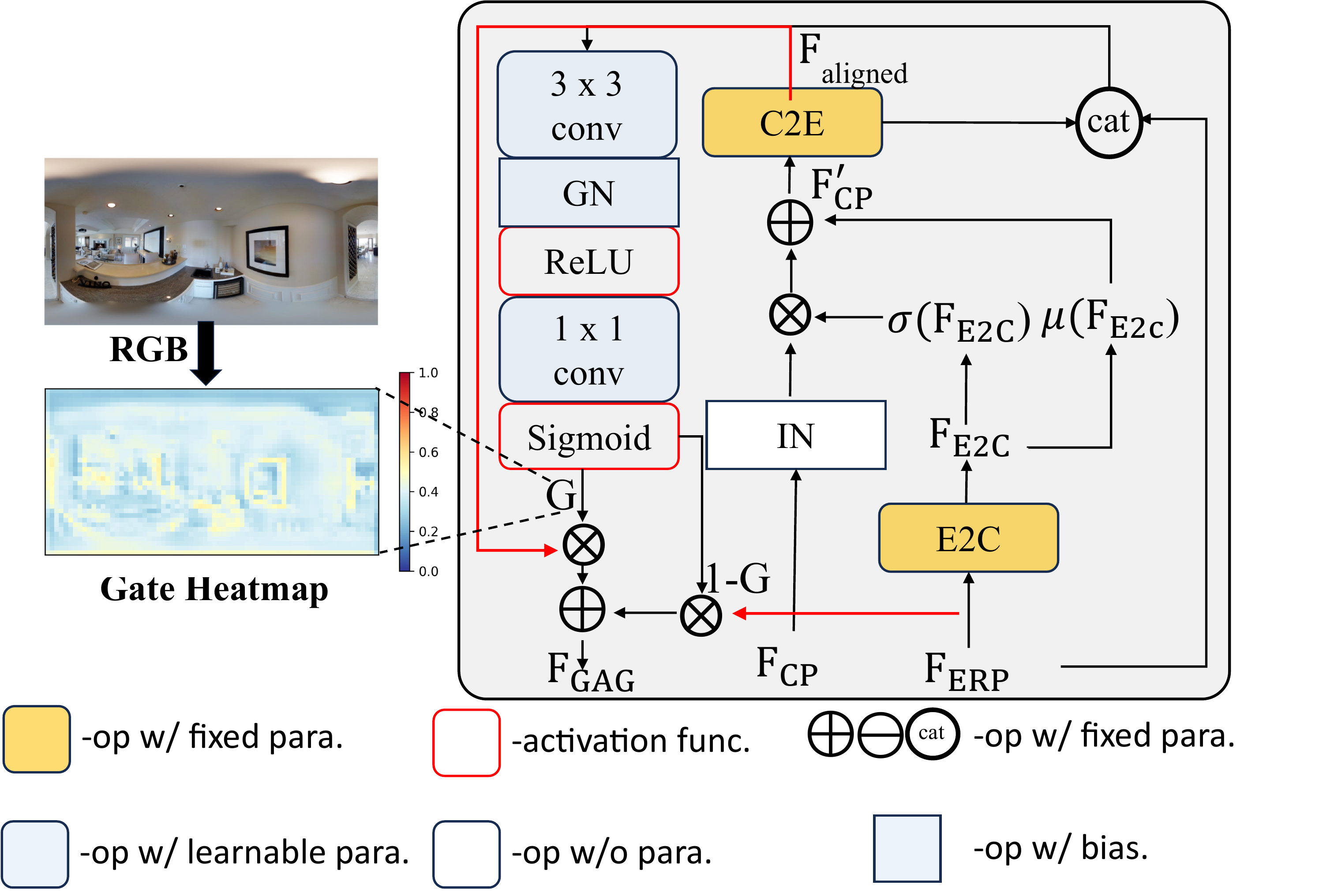}
\caption{
Geometry-Aligned Guidance (GAG).
Left: learned gate between aligned CP and ERP features, where red/blue indicates stronger CP/ERP contribution.
Right: guidance construction process, where IN denotes instance normalization, E2C ERP-to-cubemap projection, and C2E cubemap-to-ERP projection.
}
\label{fig:GAG}
\end{figure}
ERP features, in contrast, preserve the continuous panorama layout and provide stable global context, but they may contain stronger projection distortion. GAG combines these complementary features by first matching the CP feature statistics to the ERP features in the cubemap domain, and then using a learned gate to mix the aligned CP feature with the ERP feature. As shown in Fig.~\ref{fig:GAG}, IN denotes instance normalization, and the gate heatmap indicates whether each region relies more on CP or ERP information.

The guidance construction process is implemented as follows. Let $\mathbf{F}_{\text{ERP}} \in \mathbb{R}^{1 \times C \times H \times W}$ and $\mathbf{F}_{\text{CP}} \in \mathbb{R}^{6 \times C \times H_c \times W_c}$ denote the input features from the ERP and CP branches respectively. Since these two features are defined in different projection domains, we first project $\mathbf{F}_{\text{ERP}}$ into the cubemap domain, yielding $\mathbf{F}_{\text{E2C}}$ for subsequent processing.
Next, we perform parameter-free channel-wise statistical alignment between domains via affine normalization.
Specifically, we compute the channel-wise mean and standard deviation over spatial dimensions:

\begin{align}
\mu_{\text{E2C}} &= \overline{\mathbf{F}}_{\text{E2C}}^{\,HW}, &
\sigma_{\text{E2C}} &= \sqrt{\overline{\left(\mathbf{F}_{\text{E2C}}-\mu_{\text{E2C}}\right)^2}^{\,HW}+\epsilon},\\
\mu_{\text{CP}}  &= \overline{\mathbf{F}}_{\text{CP}}^{\,HW},  &
\sigma_{\text{CP}}  &= \sqrt{\overline{\left(\mathbf{F}_{\text{CP}}-\mu_{\text{CP}}\right)^2}^{\,HW}+\epsilon}.
\end{align}
Here $\mathbf{F}\in\mathbb{R}^{B\times C\times H\times W}$, and $\overline{(\cdot)}^{\,HW}$ denotes averaging over spatial locations:
\begin{align}
\overline{\mathbf{F}}^{\,HW}_{b,c} \;=\; \frac{1}{HW}\sum_{h=1}^{H}\sum_{w=1}^{W}\mathbf{F}_{b,c,h,w}.
\end{align}

This operation can be understood as first normalizing the CP feature using its own channel-wise statistics, and then rescaling it using the statistics of the ERP feature after projection to the cubemap domain. The aligned CP features are obtained by:
\begin{equation}
\mathbf{F'}_{\text{CP}} = \sigma_{\text{E2C}} \cdot \frac{\mathbf{F}_{\text{CP}} - \mu_{\text{CP}}}{\sigma_{\text{CP}}} + \mu_{\text{E2C}}.
\end{equation}
Here, $\mathbf{F}'_{\text{CP}}$ represents the CP features whose mean ($\mu_{\text{CP}}$) and standard deviation ($\sigma_{\text{CP}}$) have been aligned to match those of the ERP features ($\mu_{\text{E2C}}$ and $\sigma_{\text{E2C}}$). This alignment preserves the local structure of CP features while reducing the distribution gap between CP and ERP features.

To enable subsequent processing in the ERP domain, the calibrated cubemap features $\mathbf{F'}_{\text{CP}}$ are projected back to the original equirectangular coordinate system, producing the geometrically aligned features $\mathbf{F}_{\text{aligned}}$. After this projection, $\mathbf{F}_{\text{aligned}}$ has the same spatial layout as $\mathbf{F}_{\text{ERP}}$, so the two features can be adaptively combined.
Building on this alignment, an adaptive gating mechanism performs region-aware feature selection. The gating weights are computed as:
\begin{align}
\mathbf{G} &= \text{Sigmoid} \left( \mathcal{F}_{\text{g}} \left( [\mathbf{F}_{\text{aligned}}, \mathbf{F}_{\text{ERP}}] \right) \right).
\end{align}
where $\mathbf{G} \in [0,1]^{1 \times C \times H \times W}$ represents the spatially adaptive weighting, $\mathcal{F}_{\text{g}}$ denotes the gating function, and $[\cdot, \cdot]$ represents the concatenation operation. The gate is predicted from both the aligned CP feature and the ERP feature, allowing the model to choose the more suitable source according to local content.

The final guidance representation is constructed by adaptively aggregating the aligned CP features and the original ERP features through learned gating weights:
\begin{equation}
\mathbf{F}_{\text{GAG}} = \mathbf{G} \odot \mathbf{F}_{\text{aligned}} + (\mathbf{I} - \mathbf{G}) \odot \mathbf{F}_{\text{ERP}}.
\end{equation}
where $\odot$ denotes element-wise multiplication and $\mathbf{I}$ is an all-ones matrix with the same dimensions as $\mathbf{G}$.

Importantly, $\mathbf{F}_{\text{GAG}}$ is not used as the final backbone representation, but instead parameterizes the subsequent SCAdaLN-Zero modulation. By preserving CP-derived local structures while incorporating ERP-based global context, it provides geometry-aware guidance for structure-aware adaptation.
\subsection{Self-Conditioned AdaLN-Zero Module} 
\label{sec:SCAdaLN} 
\begin{figure}[t] 
\centering 
\includegraphics[width=0.55\textwidth]{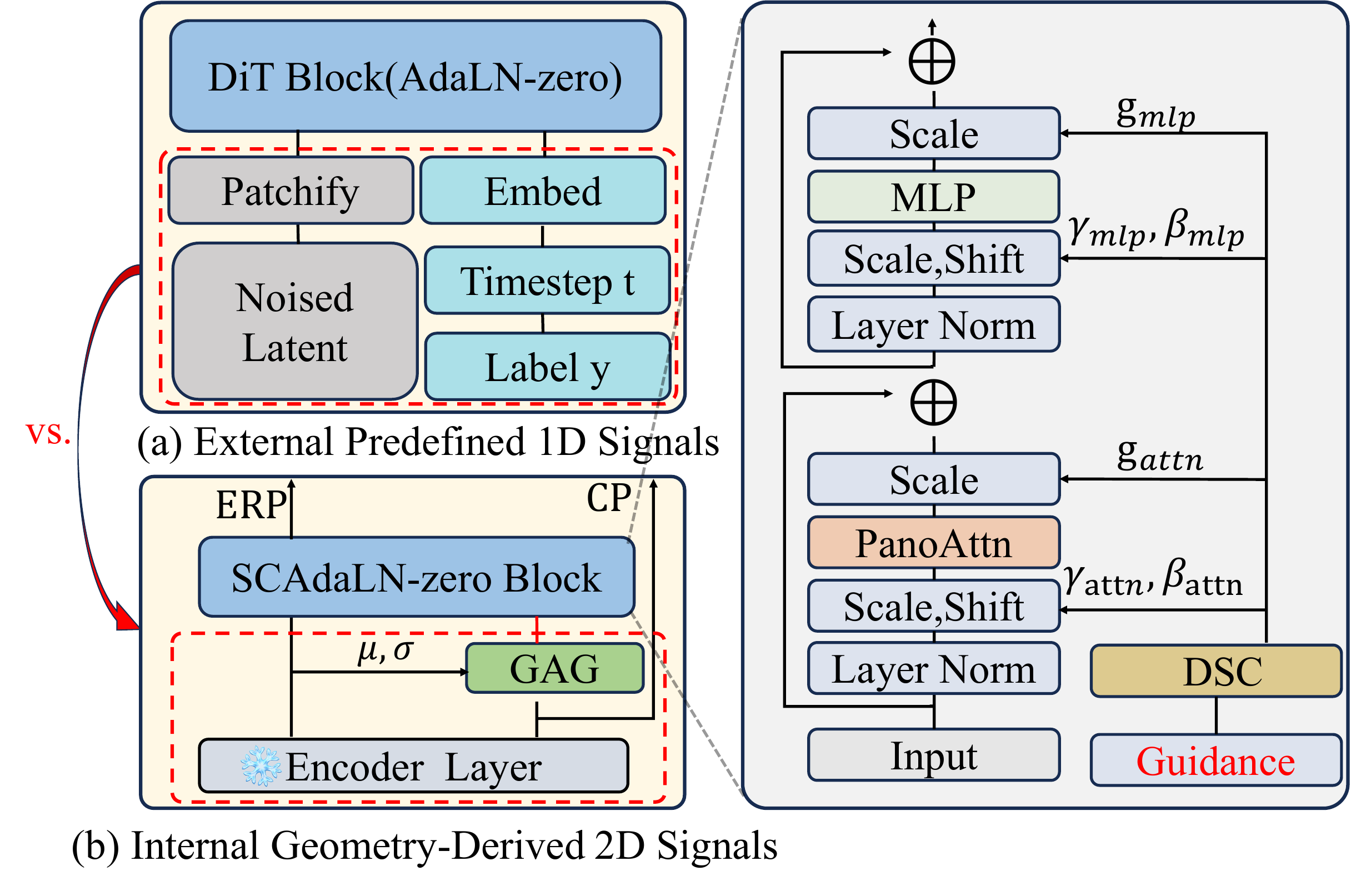} 
\caption{The top-left panel illustrates the standard DiT block. The bottom-left panel shows our SCAdaLN-Zero extension embedded in the frozen backbone. The right side details the flow: it utilizes an internal geometry-derived signal from GAG as a self-conditioning signal to modulate backbone features.} 
\label{fig:SCAdaLNZero} 
\end{figure}

We introduce a Self-Conditioned AdaLN-Zero (SCAdaLN-Zero) module to adapt ERP backbone features using the guidance feature produced by GAG. As shown in Fig.~\ref{fig:SCAdaLNZero}, unlike DiT-style AdaLN-Zero that uses external predefined 1D conditions such as timesteps or labels, our module uses an internal geometry-derived 2D signal from GAG. This guidance feature is used to predict the scale, shift, and gate parameters for the attention and MLP paths, rather than being directly fused into the backbone features.

This design is motivated by the need to adapt a perspective-pretrained backbone to panoramic inputs without disrupting its learned representation. The backbone already contains useful geometric priors from perspective image pretraining, while ERP inputs introduce projection distortion and distribution shift. A direct way to use cross-projection information is residual fusion or cross-attention. However, the ablation results in Table~\ref{tab:ablation} show that such direct integration can degrade performance, suggesting that stable adaptation depends on how the cross-projection guidance is injected. Therefore, SCAdaLN-Zero uses the GAG feature only to generate normalization parameters, avoiding direct overwrite of the backbone feature content.
Our design follows two key insights. First, scale and shift modulation provides a more controlled way to adapt pretrained features than explicit feature fusion, because it reweights normalized features instead of replacing their content. Second, the conditioning signal is generated inside the network from GAG, rather than from external prompts, labels, or timesteps. We therefore call the module self-conditioned AdaLN-Zero. With this design, the backbone can use geometry-aligned cross-projection cues to handle panoramic distortion while preserving the structure learned during pretraining.

\noindent\textbf{Modulation Parameter Generation.} Using the geometry-aware guidance signal $\mathbf{F}_{\text{GAG}} \in \mathbb{R}^{1 \times C \times H \times W}$ from the GAG module, we generate modulation parameters via a lightweight network comprising a SiLU activation followed by a depthwise separable convolution~\cite{chollet2017xception} (DSC):
\begin{equation}
\mathbf{P} = \text{DSC}(\,\text{SiLU}(\mathbf{F}_{\text{GAG}})\,).
\end{equation}
This architecture efficiently produces a parameter tensor $\mathbf{P} \in \mathbb{R}^{1 \times 6C \times H \times W}$, which is subsequently split along the channel dimension into six independent parameter groups: $\boldsymbol{\beta}_{\text{attn}}$, $\boldsymbol{\gamma}_{\text{attn}}$, $\mathbf{g}_{\text{attn}}$, $\boldsymbol{\beta}_{\text{mlp}}$, $\boldsymbol{\gamma}_{\text{mlp}}$, and $\mathbf{g}_{\text{mlp}}$, where each group corresponds to modulation parameters for different components of the Transformer block.

These parameters facilitate fine-grained, implicit modulation across two critical paths in the Transformer block via a modulation function defined as:

\begin{equation}
\text{modulate}(\mathbf{F}, \boldsymbol{\beta}, \boldsymbol{\gamma}) = \mathbf{F} \odot (1 + \boldsymbol{\gamma}) + \boldsymbol{\beta}.
\label{eq:modulate}
\end{equation}
where $\mathbf{F}$ denotes the normalized input features, $\boldsymbol{\beta}$ and $\boldsymbol{\gamma}$ represent the adaptive scaling and shifting parameters, and $\odot$ indicates element-wise multiplication.

In the self-attention path, the input features are first normalized using a LayerNorm layer configured without its standard learnable affine parameters. This design ensures that the subsequent adaptive scaling and shifting serve as the sole affine transformation after normalization. The normalized features are then transformed by the modulate function (Eq.~\ref{eq:modulate}) using parameters $\boldsymbol{\beta}_{\text{attn}}$ and $\boldsymbol{\gamma}_{\text{attn}}$, processed by a panoramic self-attention~\cite{Shen2022PanoFormerPT} (PanoAttn) layer, and finally modulated by the gating parameter $\mathbf{g}_{\text{attn}}$ to control the strength of the residual connection. In the modulated feed-forward network path, after applying LayerNorm to intermediate features and modulation via Eq.~\ref{eq:modulate} with parameters $\boldsymbol{\beta}_{\text{mlp}}$ and $\boldsymbol{\gamma}_{\text{mlp}}$, we process them through an MLP layer, and finally gate them using $\mathbf{g}_{\text{mlp}}$ to regulate the residual connection.

\noindent\textbf{Zero-Initialization Strategy.} Following DiT, we zero-initialize the final convolution in the modulation network, so the module initially degenerates to a standard Transformer block, ensuring stable training.

SCAdaLN-Zero reformulates panoramic adaptation as task-aligned parameter-space modulation rather than generic fine-tuning. By injecting geometry-aware cross-projection signals into normalization parameters, it enables distortion-aware refinement while preserving pretrained perspective priors.
\subsection{Loss Function} 
\label{sec:ECC}
To mitigate distortion-induced imbalance in ERP supervision, we introduce a cubemap-domain consistency constraint, termed the E2C Consistency Loss (ECCLoss). This design is motivated by the observation that spherical projection causes polar regions to occupy disproportionate pixel coverage in ERP, while informative geometric structures are often concentrated in equatorial areas. Such mismatch between pixel distribution and information density may bias supervision toward distortion-heavy regions and weaken depth learning in geometrically informative areas.
We therefore apply depth constraints in the CP domain, where each face follows standard perspective geometry and naturally separates polar and equatorial regions. This formulation reduces the impact of spherical distortion present in ERP supervision and provides a perspective-consistent domain for learning depth structures in a more geometrically balanced manner.

The core formulation transforms both the predicted and ground truth depth maps from ERP to CP, where $\mathbf{D}^x$ and $\mathbf{D}$ denote the predicted and ground truth depth maps in cubemap format $\mathbb{R}^{6 \times 1 \times H_c \times W_c}$, respectively. The loss function employs a Scale-Shift Invariant Mean Absolute Error~\cite{eigen2014depth} (SSI-MAE) formulation:
\begin{equation}
\mathcal{L}_{\text{ECC}} = \frac{1}{B} \sum_{i=1}^{B} \left\| \psi(\mathbf{D}_{i}) - \psi(\mathbf{D}_{i}^x) \right\|_1.
\end{equation}
Here, $B$ denotes the number of depth maps used for loss computation, typically set to $B=6$ corresponding to the six faces of the cubemap. The function $\psi(\cdot)$ represents the scale-shift normalization function:
\begin{equation}
    \psi(\mathbf{D}) = \frac{\mathbf{D} - \mu(\mathbf{D})}{\sigma(\mathbf{D}) + \epsilon}.
\end{equation}
The moments $\mu(\mathbf{D})$ and $\sigma(\mathbf{D})$ are computed over the full depth map by averaging over $(H,W)$ (not per channel):
\begin{equation}
\mu(\mathbf{D})=\frac{1}{HW}\sum_{h,w} D_{h,w},\quad
\sigma(\mathbf{D})=\sqrt{\frac{1}{HW}\sum_{h,w}\bigl(D_{h,w}-\mu(\mathbf{D})\bigr)^2+\epsilon}.
\end{equation}
with $\epsilon$ for numerical stability.

The geometric rationale of this approach stems from three advantages of CP: First, it naturally separates polar and equatorial regions through six planes, minimizing spherical distortion. Second, each face preserves perspective geometry, enabling learning of authentic 3D structure. Finally, this format reduces erroneous depth correlations between polar and equatorial regions common in ERP.

In addition to ECCLoss, our overall supervision incorporates two widely-used depth estimation losses: SILog loss~$L_{\text{SILog}}$~\cite{eigen2014depth} and gradient-domain loss~$L_{\text{Grad}}$~\cite{yin2023metric3d}. The final supervised loss function is defined as:
\begin{equation}
\begin{aligned}
L_S(\mathbf{D}^x, \mathbf{D}) &= L_{\text{SILog}}(\mathbf{D}^x, \mathbf{D}) + L_{\text{Grad}}(\mathbf{D}^x, \mathbf{D}) \\
&\quad + \lambda_E \cdot L_{\text{ECC}}(\mathbf{D}^x, \mathbf{D}).
\end{aligned}
\end{equation}
where $\mathbf{D}^x$ and $\mathbf{D}$ denote the predicted and ground truth depth maps, with $\lambda_E$ balancing the ECCLoss contribution.

\section{Experiment}
\label{section4}
\subsection{Implementation Details}

\noindent\textbf{Datasets.} We employ two real-world indoor panoramic datasets, Matterport3D~\cite{chang2017matterport3d} and Stanford2D3D~\cite{Armeni2017Joint2D}, for evaluation under two settings: for in-domain evaluation, models are trained directly on these two datasets; for zero-shot evaluation, models are trained exclusively on synthetic data from Structured3D~\cite{Zheng2019Structured3DAL} and Deep360~\cite{li2022mode}. Following PanDA's evaluation protocol, we train at $504\times 1008$ and upsample all predictions to $512\times 1024$ for evaluation. 

\noindent\textbf{Implementation Details.} All experiments are performed on eight NVIDIA RTX A4000 GPUs using the PyTorch framework. We employ the Adam optimizer~\cite{kingma2014adam} with an initial learning rate of $5\times 10^{-5}$, while keeping other hyperparameters at their default values. During training, we use a batch size of 1 and adopt several data augmentation strategies, including left-right flipping, panoramic horizontal rotation, and luminance adjustment, following established practices in~\cite{Jiang2021UniFuseUF, Shen2022PanoFormerPT}. Our backbone network is the frozen Depth Anything Model v2~\cite{DAMv2} (DAM v2), initialized with the same pretrained weights as PanDA~\cite{cao2025panda}. To balance efficiency and computational cost, we integrate our proposed modules into the odd-numbered layers, with layer ablations provided in the supplementary material.

\subsection{Comparison Analysis}
\begin{table*}[t]
\centering
\scriptsize
\setlength{\tabcolsep}{2.5pt}
\caption{
Quantitative comparison with SOTA methods on Matterport3D and Stanford2D3D datasets.
The best results are in \textbf{bold}, and the second best are \underline{underlined}.
The $\Delta$ row indicates the relative improvement (in percentage) of our method
over the previous best method (PanDA-L) under fair comparison settings.
Note that the training protocols differ:
PanDA-L* was first pre-trained using semi-supervised learning on 120K panoramic images
and then fine-tuned on in-domain data,
whereas our method and all other baselines were trained directly on in-domain data only.
}
\begin{tabular}{llcccccc}
\toprule
\textbf{Dataset} & \textbf{Method} 
& Abs Rel $\downarrow$
& Sq Rel $\downarrow$
& RMSE $\downarrow$
& $\delta_1$ $\uparrow$
& $\delta_2$ $\uparrow$
& $\delta_3$ $\uparrow$ \\
\midrule

\multirow{12}{*}{Matterport3D~\cite{chang2017matterport3d}}
& BiFuse~\cite{Wang2020BiFuseM3} & 0.2048 & - & 0.6259 & 84.52 & 93.19 & 96.32 \\
& UniFuse~\cite{Jiang2021UniFuseUF} & 0.1063 & - & 0.4941 & 88.97 & 96.23 & 98.31 \\
& BiFuse++~\cite{Wang2022BiFuseSA} & - & - & 0.5190 & 87.90 & 95.17 & 97.72 \\
& PanoFormer~\cite{Shen2022PanoFormerPT} & - & - & 0.3635 & 91.84 & 98.04 & 99.16 \\
& HRDFuse~\cite{Ai2023HRDFuseM3} & 0.0967 & 0.0936 & 0.4433 & 91.62 & 96.69 & 98.44 \\
& Elite360D~\cite{Ai_2024_CVPR} & 0.1115 & 0.0914 & 0.4875 & 88.15 & 96.46 & 98.74 \\
& Depth Anywhere~\cite{wang2024depth} & - & 0.0850 & - & 91.70 & 97.60 & 99.10 \\
& SGFormer~\cite{zhang2025sgformer} & 0.1039 & 0.0865 & 0.4790 & 89.46 & 96.42 & 98.59 \\
& PanDA-L*~\cite{cao2025panda} & \underline{0.0717} & - & \underline{0.3305} & \underline{95.09} & \underline{98.94} & \underline{99.65} \\
\rowcolor{gray!13} & PanDA-L & 0.0788 & \underline{0.0572} & 0.3827 & 93.73 & 97.90 & 99.44 \\
\rowcolor{gray!13} &RePer-360 (Ours) 
& \textbf{0.0691} & \textbf{0.0388} & \textbf{0.3164} 
& \textbf{95.67} & \textbf{99.07} & \textbf{99.69} \\
\rowcolor{gray!13} &$\Delta$ 
& 12.3\% & 32.2\% & 17.3\% 
& 2.07\% & 1.19\% & 0.25\% \\

\midrule

\multirow{12}{*}{Stanford2D3D~\cite{Armeni2017Joint2D}}
& BiFuse~\cite{Wang2020BiFuseM3} & 0.1209 & - & 0.4142 & 86.60 & 95.80 & 98.60 \\
& UniFuse~\cite{Jiang2021UniFuseUF} & 0.1114 & - & 0.3691 & 87.11 & 96.64 & 98.82 \\
& BiFuse++~\cite{Wang2022BiFuseSA} & - & 0.3720 & - & 87.83 & 96.49 & 98.84 \\
& PanoFormer~\cite{Shen2022PanoFormerPT} & - & 0.3083 & - & 93.94 & 98.38 & 99.41 \\
& HRDFuse~\cite{Ai2023HRDFuseM3} & 0.0935 & 0.0508 & 0.3106 & 91.40 & 97.98 & 99.27 \\
& Elite360D~\cite{Ai_2024_CVPR} & 0.1182 & 0.0728 & 0.3756 & 88.72 & 96.84 & 98.92 \\
& Depth Anywhere~\cite{wang2024depth} & 0.1180 & - & 0.3510 & 91.00 & 97.10 & 98.70 \\
& SGFormer~\cite{zhang2025sgformer} & 0.1040 & 0.0581 & 0.3406 & 89.98 & 96.93 & 99.08 \\
& PanDA-L*~\cite{cao2025panda} & \underline{0.0609} & - & \underline{0.2540} & \underline{96.82} & \textbf{99.05} & \textbf{99.52} \\
\rowcolor{gray!13} & PanDA-L & 0.0881 & \underline{0.0550} & 0.3185 & 93.42 & 98.26 & 99.33 \\
\rowcolor{gray!13} &RePer-360 (Ours) 
& \textbf{0.0580} & \textbf{0.0382} & \textbf{0.2474} 
& \textbf{97.37} & \underline{98.89} & \underline{99.35} \\
\rowcolor{gray!13}& $\Delta$ 
& 34.2\% & 30.5\% & 22.3\% 
& 4.22\% & 0.64\% & 0.02\% \\

\bottomrule
\end{tabular}
\label{tab:quanti}
\end{table*}
\begin{table}[t]
\centering
\scriptsize
\setlength{\tabcolsep}{2pt}
\caption{Comparison of zero-shot depth estimation performance. The best results are in \textbf{bold}.}
\begin{tabular}{l|c|c c|c c}
\toprule 
Method & Backbone & \multicolumn{2}{c|}{Matterport3D~\cite{chang2017matterport3d}} & \multicolumn{2}{c}{Stanford2D3D~\cite{Armeni2017Joint2D}} \\
& & Abs Rel $\downarrow$ & RMSE $\downarrow$ & Abs Rel $\downarrow$ & RMSE $\downarrow$ \\
\midrule
Marigold~\cite{ke2024repurposing} & SD 2.0~\cite{rombach2022high} & - & 0.5745 & - & 0.5069 \\
\midrule
DAM v1~\cite{DAMv2} & ViT-L & - &1.1431 & - & 0.7597\\
DAM v2~\cite{DAMv2} & ViT-L & - & 0.5522 & - & 0.4884 \\
PanDA-L~\cite{cao2025panda} & ViT-L & 0.1036 & 0.4539 & 0.1092 & 0.3314 \\
RePer-360 (Ours)& ViT-L & \textbf{0.1033} & \textbf{0.4534} & \textbf{0.0630} & \textbf{0.2849} \\
\bottomrule 
\end{tabular}
\label{tab:ZeroShot}
\end{table}
\begin{figure*}[t]
  \centering
\includegraphics[width=0.9\textwidth]{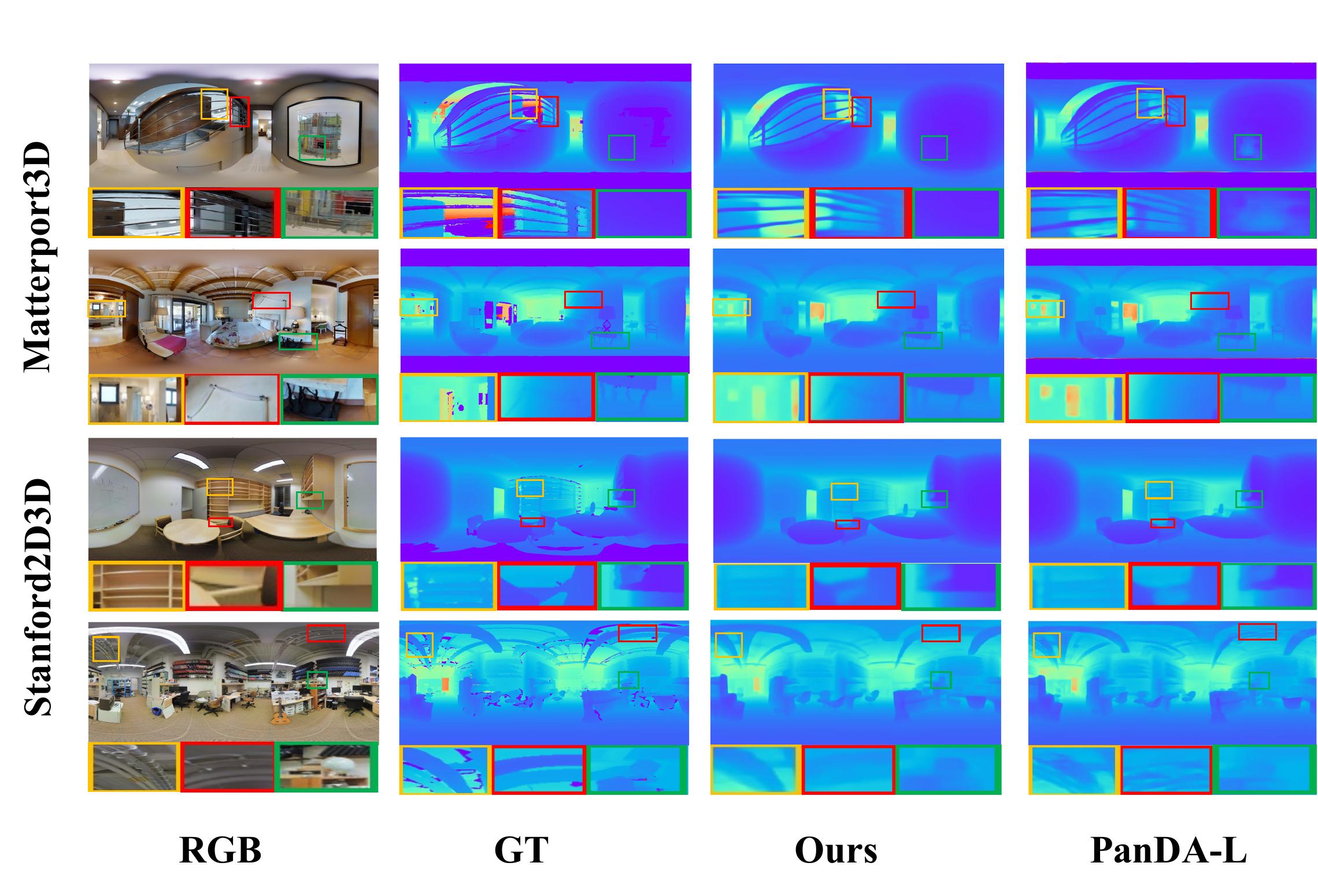}
  \caption{Qualitative comparison between the current SOTA method PanDA-L~\cite{cao2025panda} and ours. The results of PanDA-L are obtained using their released weights. The top sample is from Matterport3D, while the bottom one is from Stanford2D3D.}
  \label{fig:qualitative}
\end{figure*}

\begin{figure*}[t]
\centering
\includegraphics[width=0.8\textwidth]{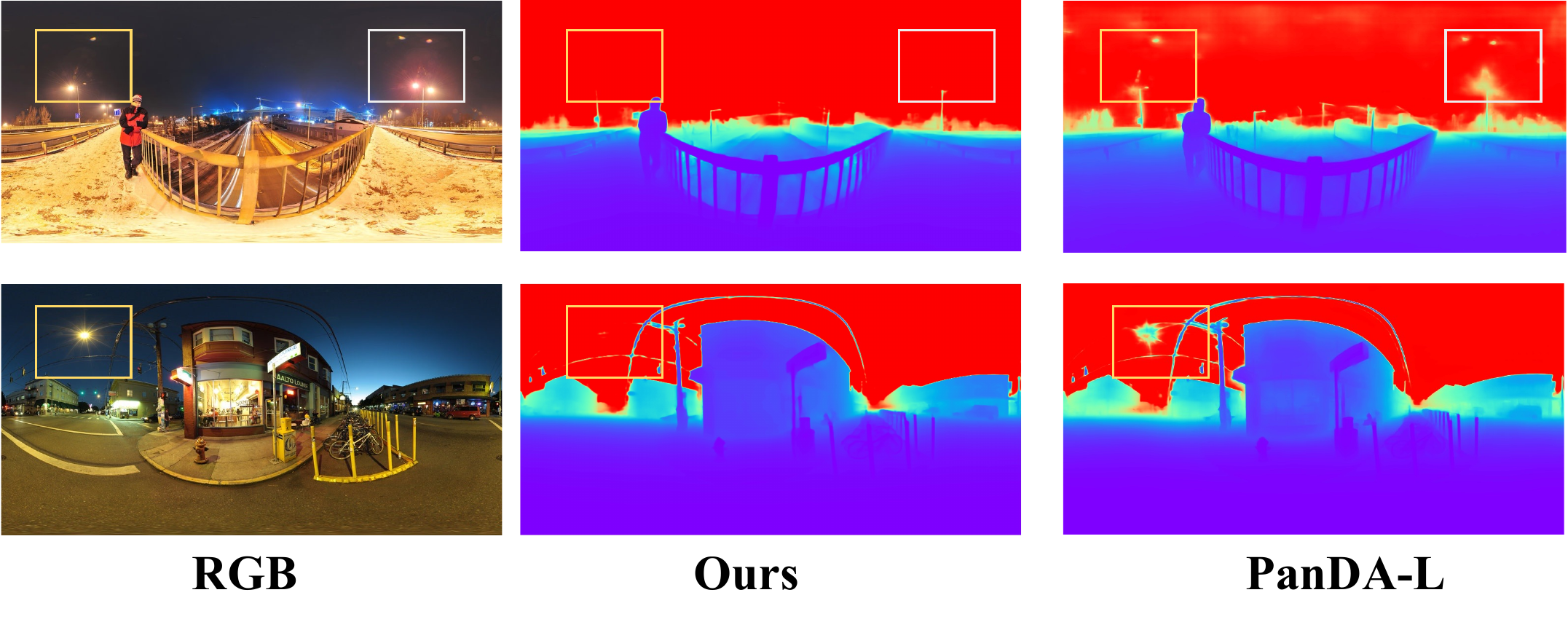}
  \caption{Zero-shot qualitative comparison on the unlabeled SUN360~\cite{xiao2012recognizing} dataset.
  Our model is trained only on Structured3D~\cite{Zheng2019Structured3DAL} and Deep360~\cite{li2022mode},
  while PanDA-L~\cite{cao2025panda} uses substantially more panoramic supervision.
  Despite the limited training data, our method better preserves structural geometry under complex outdoor illumination.}
  \label{fig:qualitative_outdoor}
\end{figure*}
\noindent\textbf{Quantitative Results.} As shown in Table~\ref{tab:quanti}, our method achieves SOTA performance on both Matterport3D~\cite{chang2017matterport3d} and Stanford2D3D~\cite{Armeni2017Joint2D} datasets. Notably, PanDA-L~\cite{cao2025panda} first undergoes large-scale semi-supervised multi-dataset pretraining
on four panoramic datasets (Structured3D~\cite{Zheng2019Structured3DAL}, Deep360~\cite{li2022mode}, ZInD~\cite{cruz2021zillow},
and 360+x~\cite{chen2024360+}), totaling about 120K samples, and is then fine-tuned on in-domain data.
In contrast, our method surpasses it using only in-domain training data (1K for Stanford2D3D and 8K for Matterport3D),
highlighting data efficiency. In a fair comparison with the non-pretrained PanDA-L, our advantages become more pronounced: improving Abs Rel by 12.3\% and RMSE by 17.3\% on Matterport3D, while achieving 34.2\% and 22.3\% improvements respectively on Stanford2D3D. The zero-shot generalization results in Table~\ref{tab:ZeroShot} further validate our advantages under the same synthetic-data training regime (Structured3D and Deep360 only), with improvements of 42.3\% in Abs Rel and 14.0\% in RMSE on Stanford2D3D. These results demonstrate cross-domain generalization and data efficiency. See the supplementary material for complexity analysis and comparison with ST$^2$360D.

 \noindent\textbf{Qualitative Results.} As shown in Fig.~\ref{fig:qualitative}, our method consistently outperforms PanDA-L~\cite{cao2025panda}. PanDA-L tends to misinterpret wall textures as depth variations and loses structural details in complex regions, whereas our approach preserves scene geometry and fine structures more faithfully, especially under severe panoramic distortions. These visual improvements align with our superior RMSE results, demonstrating effective distortion handling despite using substantially less training data than PanDA-L.
Fig.~\ref{fig:qualitative_outdoor} further shows zero-shot results on SUN360. Trained only on synthetic panoramas (Structured3D and Deep360), our model achieves stable geometric reconstruction under challenging outdoor illumination, while PanDA-L is more sensitive to lighting changes and occasionally confuses them with depth variations. These results highlight the strong cross-domain generalization of our method with limited panoramic data. Additional results on self-collected real-world panoramas are provided in the supplementary material.
\subsection{Architectural Comparison and Ablation}
\label{sec:ablation}

\begin{figure*}[t]
\centering
     \includegraphics[width=\textwidth]{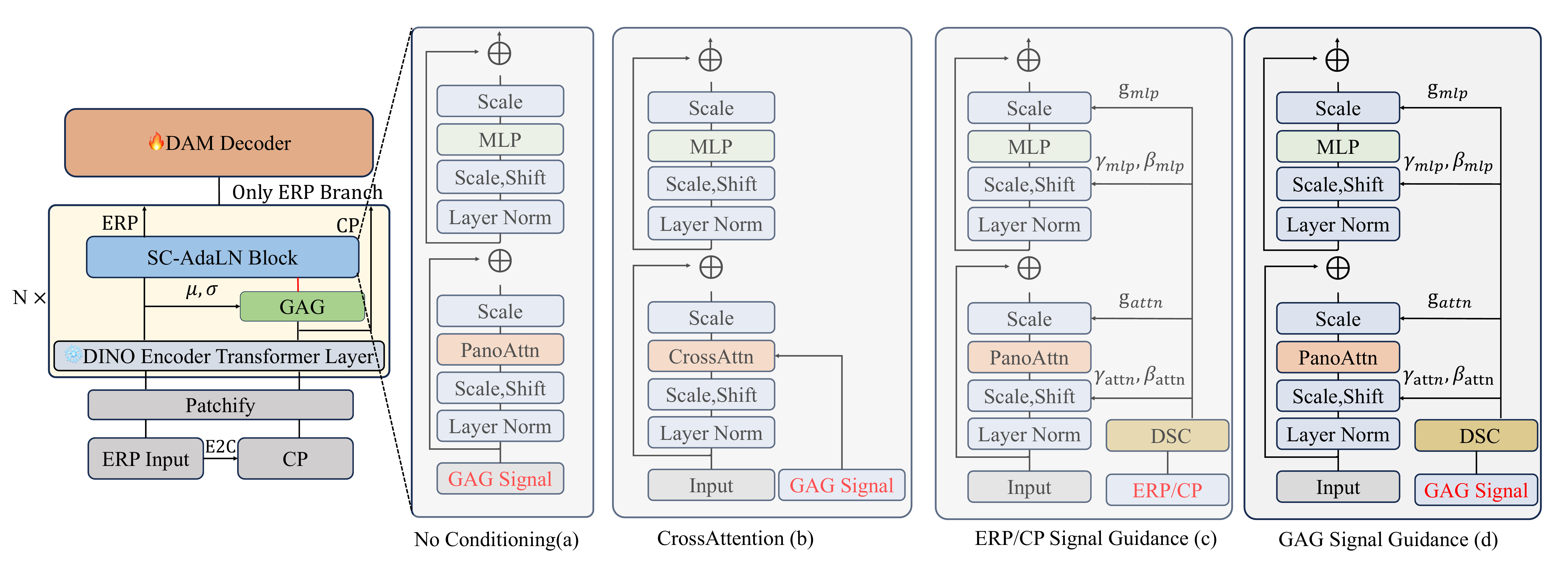} 
    \caption{
        \textbf{Architectural Ablation Study.}
        We compare different structural formulations to validate our design:
(a) a degenerated version without conditioning;
(b) replacing normalization-based modulation with explicit feature mixing via cross-attention;
(c) conditioning on a single-branch signal (ERP or CP);
and (d) the complete geometry-aware conditioning design (ours), which achieves the best performance.
    }
  \label{fig:Architectures}
  
\end{figure*}

\begin{table*}[t]
\centering
\scriptsize
\setlength{\tabcolsep}{1.5pt}
\renewcommand{\arraystretch}{0.95}
\caption{\textbf{Architectural Ablation Results.}
Quantitative comparison of different architectural variants of the proposed framework,
including alternative formulations of the SCAdaLN-based conditioning mechanism,
attention types, and the effect of ECCLoss.
Configurations (a–d) correspond to Fig.~\ref{fig:Architectures}.
Best results are shown in \textbf{bold}.}
\begin{tabular}{ccccccccccc}
\toprule 
Index & SCAdaLN 
& Guidance Source 
& Attention 
& ECCLoss 
& RMSE$\downarrow$ 
& Abs Rel$\downarrow$ 
& Sq Rel$\downarrow$
& $\delta_{1}(\%)\uparrow$ 
& $\delta_{2}(\%)\uparrow$ \\
\midrule
- & \ding{55} & None & None & \ding{55}
& 0.3846 & 0.0915 & 0.0525 & 92.25 & 98.53 \\
\midrule
a & \ding{55} & GAG & PanoAttn & \checkmark
& 0.4638 & 0.1158 & 0.0790 & 86.80 & 96.80 \\
b & \ding{55} & GAG& CrossAttn & \checkmark
& 0.4549 & 0.1168 & 0.0763 & 87.15 & 97.17 \\
c & \checkmark & ERP  & PanoAttn & \checkmark
& 0.3206 & 0.0698 & 0.0390 & 95.66 & 99.03 \\
c & \checkmark & CP  & PanoAttn & \checkmark
& 0.3227 & 0.0692 & \textbf{0.0388} & 95.58 & 99.03 \\
d & \checkmark & GAG & PanoAttn & \checkmark
& \textbf{0.3164} & \textbf{0.0691} & \textbf{0.0388} 
& \textbf{95.67} & \textbf{99.07} \\
\midrule
- & \checkmark & GAG & PanoAttn & \ding{55}
& 0.3249 & 0.0716 & 0.0396 & 95.17 & 99.04 \\
- & \checkmark & GAG & NormalAttn & \checkmark
& 0.3209 & 0.0725 & 0.0393 & 95.43 & 99.01 \\
- & \checkmark & UniFuse & PanoAttn & \checkmark
& 0.3244 & 0.0731 & 0.0412 & 94.93 & 99.04 \\
- & \ding{55} & UniFuse & None &  \checkmark
& 0.6032 & 0.1641 & 0.1426 & 78.30 & 93.73 \\
\bottomrule 
\end{tabular}
\label{tab:ablation}
\end{table*}

\begin{table}[t]
\centering
\scriptsize
\renewcommand{\arraystretch}{1.0}
\setlength{\tabcolsep}{6pt}
\caption{Impact of Pretrained Initialization on Matterport3D (ViT-L backbone).}
\begin{tabular}{lccc}
\toprule
Method & AbsRel $\downarrow$ & RMSE $\downarrow$ & $\delta_1(\%) \uparrow$ \\
\midrule
PanDA (DAMv2) & 0.0788 & 0.3827 & 93.73 \\
Ours (DAMv2)  & \textbf{0.0691} & \textbf{0.3164} & \textbf{95.67} \\
Ours (DINOv2) & 0.0808 & 0.3518 & 94.33 \\
\bottomrule
\end{tabular}
\label{tab:weights}
\end{table}
We conduct comprehensive architectural analyses on Matterport3D~\cite{chang2017matterport3d} under a unified training configuration. 
All variants share the same frozen backbone, training data, and optimization strategy to ensure fairness. 
In addition to internal structural ablations, we further compare with representative traditional panoramic fusion designs to evaluate the effectiveness of geometry-aware modulation.
Results are summarized in Table~\ref{tab:ablation}.

\noindent\textbf{Effectiveness of SCAdaLN-Zero.}
As shown in Table~\ref{tab:ablation} (Index a–d), replacing explicit interaction (Index a/b) with SCAdaLN-Zero (Index c/d) yields clear improvements. 
Even single-branch conditioning (ERP or CP) outperforms the degenerated baseline, while the full geometry-aligned configuration (Index d) performs best. 
These results show that normalization-based modulation is more effective than explicit value-level feature mixing for cross-domain adaptation.

\noindent\textbf{Effect of Geometry-Aligned Guidance.}
Comparing Index c (ERP/CP single-branch variants) with Index d, incorporating GAG conditioning further improves performance over single-branch signals.
This indicates that geometry-aligned cross-projection features provide more informative and consistent modulation cues than using isolated ERP or CP features alone.

\noindent\textbf{Fusion Architectural Variants.}
We evaluate two variants using the CEE module from UniFuse:
replacing GAG with CEE fusion while retaining SCAdaLN,
and adopting a full fusion design without modulation.
The former underperforms geometry-aligned conditioning,
while the latter incurs a substantial drop,
indicating that multi-branch fusion is insufficient
for projection-gap adaptation.

\noindent\textbf{Role of ECCLoss and Attention Design.}
As shown in the lower rows of Table~\ref{tab:ablation}, removing ECCLoss leads to consistent performance degradation,
highlighting its importance in enforcing cross-projection geometric consistency.
Replacing panoramic attention with standard self-attention produces only minor performance differences,
indicating that the proposed framework is not tightly coupled to a specific attention operator.

\noindent\textbf{Impact of Pretrained Initialization.}
We evaluate the role of pretrained weights using the same ViT-L architecture (Table~\ref{tab:weights}).
With identical DAMv2 initialization, our method consistently outperforms the LoRA-based PanDA baseline.
When initialized from task-agnostic DINOv2 weights, RePer-360 remains competitive,
suggesting that the proposed adaptation is not solely dependent on depth-pretrained initialization.
\subsection{Representation Drift and Feature Visualization}
\label{sec:analysis}
To better understand different adaptation mechanisms, we analyze representation drift and feature-level responses.

\begin{figure*}[t]
\centering
\includegraphics[width=\textwidth]{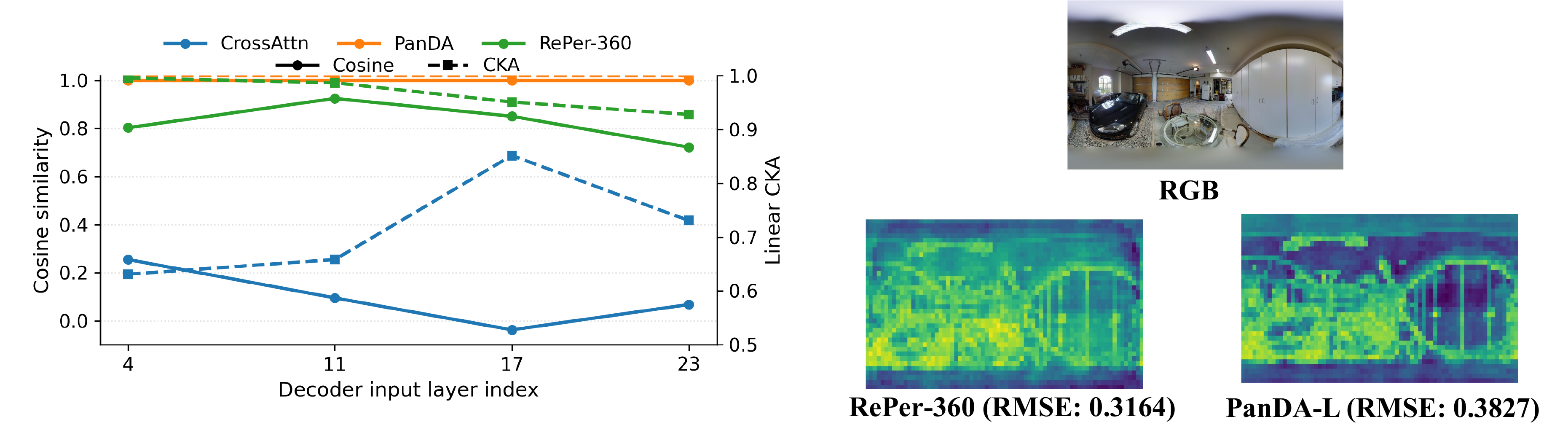}
\caption{
    \textbf{Left:} Feature drift relative to the frozen backbone on Matterport3D
    (the first 500 test images), measured using cosine similarity and linear CKA
    across transformer layers.
    \textbf{Right:} Visualization of layer-4 (decoder-input) features for an example image,
    comparing RePer-360 and the in-domain PanDA baseline.
    Both methods capture a similar near-to-far transition, while PanDA shows over-activation on ceiling/lighting regions and an exaggerated ceiling--wall contrast.
}
\label{fig:analysis}
\end{figure*}

\noindent\textbf{Representation Drift.}
The left panel of Fig.~\ref{fig:analysis} reports feature similarity to the frozen backbone
on Matterport3D using cosine similarity and linear CKA.
Cross-attention induces large representational shifts with unstable layer-wise trends.
In contrast, PanDA stays close to the backbone, while RePer-360 preserves high similarity with smooth, consistent inter-layer evolution,
indicating controlled, geometry-aware adaptation rather than layer-wise instability.
Overall, effective panoramic adaptation benefits from balanced representation reconfiguration instead of excessive perturbation or near-identity preservation.

\noindent\textbf{Feature Map Visualization.}
We visualize decoder-input features at layer 4 to compare PanDA and RePer-360.
As shown in Fig.~\ref{fig:analysis}, both methods exhibit a consistent near-to-far intensity transition,
indicating that the depth/layout trend is captured.
However, PanDA shows disproportionately strong responses on ceiling regions and lighting fixtures,
resulting in an exaggerated ceiling--wall contrast.
In contrast, RePer-360 maintains a more coherent response on these planar regions,
consistent with the controlled drift trends in the left panel.

\section{Conclusion}
We introduce RePer-360 to bridge the projection gap between perspective and panoramic imagery through geometry-aware, self-conditioned normalization-based modulation. By preserving and reorganizing pretrained perspective priors rather than overwriting them via feature fusion, our framework achieves structured cross-domain adaptation with strong data efficiency. These findings suggest that balanced prior modulation offers a principled strategy for adapting pretrained visual backbones to geometrically mismatched domains.

\paragraph{\textbf{Acknowledgement}. This work was supported by the National Natural Science Foundation of China (Nos. 62573039, U2441242).}

\bibliographystyle{splncs04}
\bibliography{main}

\clearpage
\begin{center}
{\LARGE \textbf{RePer-360: Releasing Perspective Priors for 360$^\circ$ Depth Estimation via Self-Modulation}}\\[10pt]
{\Large \textbf{Supplementary Material}}
\end{center}

\appendix
\setcounter{section}{0} 
\renewcommand\thesection{\Alph{section}}

\section{Analysis}
\label{sec:analysis}

\subsection{Equirectangular to Cubemap Transformation}

The transformation from an Equirectangular Projection (ERP) to a Cubemap Projection (CP) is a geometric re-projection process that maps spherical data onto the six faces of a unit cube. Let $I_{ERP}$ denote the source equirectangular image with resolution $W \times H$. The target cubemap consists of six square faces $\{F_{front}, F_{back}, F_{left}, F_{right}, F_{up}, F_{down}\}$, each with side length $H/2$.

For each pixel $(u,v)$ on a cubemap face, the corresponding pixel value in $I_{ERP}$ is obtained through inverse mapping. First, the 2D pixel coordinate $(u,v)$ on the cube face is converted into a normalized 3D direction vector $\mathbf{v}=(x,y,z)$ in Cartesian space. This direction vector is then converted into spherical coordinates $(\phi,\theta)$:

\begin{align}
\phi &= \arctan2(z, x),\\
\theta &= \arcsin\left(\frac{y}{\sqrt{x^2+y^2+z^2}}\right)
\end{align}

where $\phi \in [-\pi,\pi]$ and $\theta \in [-\frac{\pi}{2},\frac{\pi}{2}]$. The spherical coordinates are subsequently mapped to normalized UV coordinates $(u',v')$ in the ERP image:

\begin{align}
u' &= \frac{\phi}{2\pi}+0.5,\\
v' &= \frac{\theta}{\pi}+0.5
\end{align}

Finally, bilinear interpolation is applied at coordinates $(u'W, v'H)$ to sample pixel values from $I_{ERP}$, ensuring smooth spatial transitions.

\begin{figure}[t]
\centering
\includegraphics[width=\textwidth]{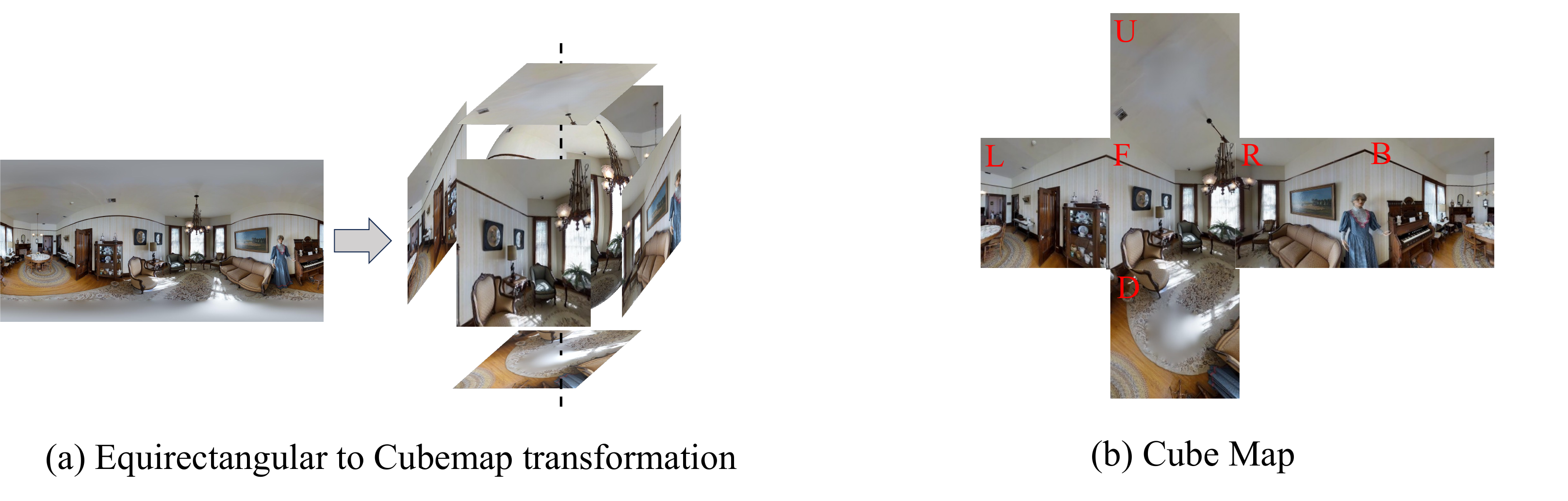}
\caption{ERP-to-cubemap transformation. (a) Projection from the spherical ERP domain onto a cube. (b) The unfolded cubemap layout with face orientations: Front (F), Back (B), Left (L), Right (R), Up (U), and Down (D).}
\label{fig:erp_to_cp}
\end{figure}

Fig.~\ref{fig:erp_to_cp} illustrates this transformation process. The spherical panorama is first projected onto a 3D cube geometry, and the resulting cubemap faces can then be unfolded into a planar layout while preserving their spatial orientation.

\subsection{Key Finding: Projection Trade-off}

Our analysis reveals that the input projection format significantly affects the performance of perspective-trained depth estimation models. Following the observations of PanDA~\cite{cao2025panda}, Cubemap Projection (CP) preserves more accurate local geometry because each face follows perspective camera geometry consistent with the model's training domain. In contrast, Equirectangular Projection (ERP) maintains global structural continuity but often suffers from degraded local detail due to severe spherical distortion.

\begin{figure}[h]
\centering
\includegraphics[width=0.8\textwidth]{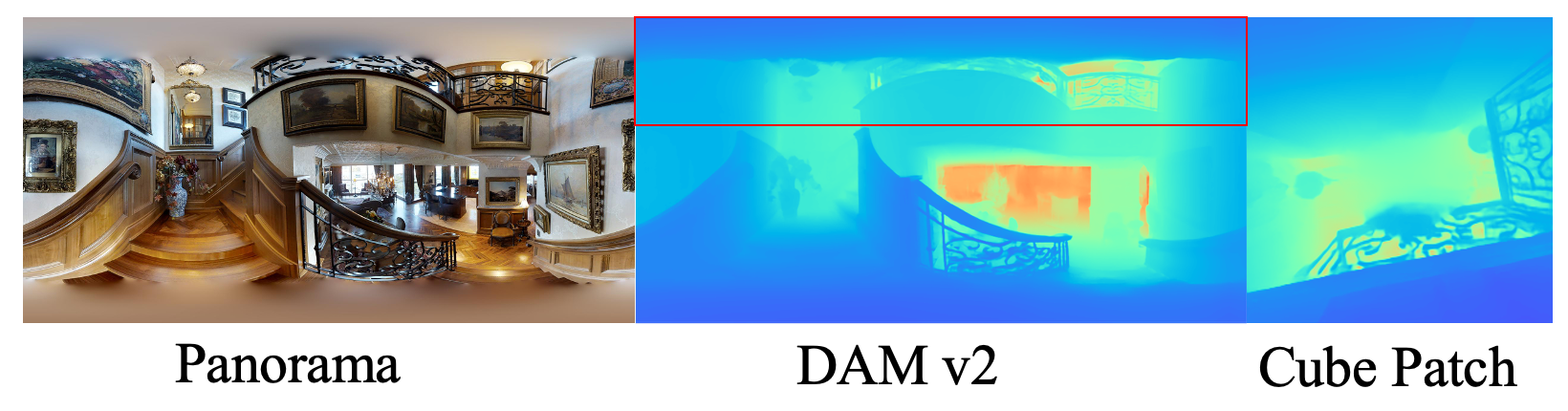}
\caption{\textbf{Projection format comparison.} ERP preserves global structure but tends to lose local detail, whereas CP improves local geometric accuracy. This trade-off motivates the design of our framework.}
\label{fig:key_finding}
\end{figure}

\noindent\textbf{ERP Limitation: Structural Integrity but Detail Loss.}
As illustrated in Fig.~\ref{fig:key_finding}, when the entire ERP image is directly processed by a perspective-trained depth estimator (\textit{e.g.}, DAM v2), the predicted depth map generally preserves the overall scene structure but lacks fine geometric details. This limitation becomes particularly evident in high-distortion regions such as ceilings, where features appear blurred or inconsistent. The reason is that ERP sampling violates the implicit perspective assumption embedded in the backbone architecture, making it difficult for the model to extract high-frequency geometric cues reliably.

\noindent\textbf{CP Advantage: Accurate Local Geometry.}
In contrast, when cubemap faces, each approximating a perspective image, are used as input, the model can recover sharper and more accurate local depth structures. The nearly uniform sampling density within each cubemap face aligns well with the perspective geometry assumed during pretraining, enabling the network to better capture fine geometric patterns.

\noindent\textbf{Design Implication: Detail-Guided Global Structure.}
These observations reveal an inherent trade-off: ERP provides globally coherent scene structure but sacrifices local detail, whereas CP captures precise local geometry but lacks global continuity and may introduce stitching artifacts. 

Our RePer-360 framework is built upon this insight. Instead of fusing ERP and CP features, the Geometry-Aligned Guidance (GAG) module extracts geometry-aware cues from CP and uses them to modulate ERP representations via the Self-Conditioned AdaLN-Zero (SCAdaLN-Zero) mechanism. This design enables CP-derived details to refine ERP features while preserving the global structural consistency of the panoramic representation.
\section{More Experimental Results}
\label{ExResults}

\subsection{Layer Ablation for Modulation}
\label{sec:layer_ablation}

We further analyze which layers should be modulated in RePer-360. The output-related layers, denoted as Out., correspond to layers 4, 11, 17, and 23. As shown in Table~\ref{tab:layer_ablation}, modulating only the output-related layers is insufficient. Adding even layers improves results, while adding odd layers achieves the best performance. In contrast, modulating all layers does not further improve performance, indicating that dense modulation is not necessarily beneficial. These results support our default Out.+Odd design, which provides a stable accuracy-efficiency trade-off.

\begin{table}[h]
\centering
\footnotesize
\setlength{\tabcolsep}{10pt}
\caption{\textbf{Layer Ablation.} Out. denotes output-related layers: layers 4, 11, 17, and 23. Out.+Odd achieves the best performance.}
\label{tab:layer_ablation}
\begin{tabular}{l |c |c}
\toprule
Layers & Abs Rel $\downarrow$ & RMSE $\downarrow$ \\
\midrule
Out. only & 0.0750 & 0.3366 \\
Out.+Even & 0.0719 & 0.3216 \\
Out.+Odd & \textbf{0.0691} & \textbf{0.3164} \\
All & 0.0733 & 0.3232 \\
\bottomrule
\end{tabular}
\end{table}

\subsection{Various Backbones}
\label{sec:backbones}

In addition to the ViT-Large (ViT-L) backbone used in our main experiments, we further evaluate the scalability of RePer-360 using smaller backbones, namely ViT-Base (ViT-B) and ViT-Small (ViT-S). Quantitative comparisons with the state-of-the-art method PanDA~\cite{cao2025panda} are reported in Table~\ref{tab:Complex}. Two notable observations emerge from these results.

\noindent\textbf{Performance Behavior under Direct Fine-tuning.}
Interestingly, we observe that the PanDA baseline with a ViT-B backbone (Abs Rel 0.0773 on Matterport3D) slightly outperforms its ViT-L counterpart reported in the main text (Abs Rel 0.0788). This counterintuitive behavior may be partially attributed to overfitting. The ViT-L model, with its larger parameter space (300M+ parameters), typically requires a larger amount of diverse training data to achieve optimal generalization. When fine-tuned directly on relatively limited in-domain panoramic datasets (Matterport3D/Stanford2D3D) without specialized adaptation mechanisms, the larger model may become more sensitive to dataset-specific noise or distortion patterns. In contrast, ViT-B provides a more balanced trade-off between model capacity and generalization under this data scale.

\begin{table}[t]
\centering
\caption{Comparison of depth estimation performance across different backbones. The best results are shown in \textbf{bold}.}
\footnotesize
\setlength{\tabcolsep}{2pt}
\begin{tabular}{l|c|c c|c c}
\toprule 
Method & Backbone & \multicolumn{2}{c|}{Matterport3D~\cite{chang2017matterport3d}} & \multicolumn{2}{c}{Stanford2D3D~\cite{Armeni2017Joint2D}} \\
& & Abs Rel $\downarrow$ & RMSE $\downarrow$ & Abs Rel $\downarrow$ & RMSE $\downarrow$ \\
\midrule
PanDA-S~\cite{cao2025panda} & ViT-S & 0.0884 & 0.4160 & 0.0865 & 0.3219 \\
RePer-360 (Ours) & ViT-S & \textbf{0.0875} & \textbf{0.4073} & \textbf{0.0801} & \textbf{0.3130} \\
\midrule
PanDA-B~\cite{cao2025panda} & ViT-B & \textbf{0.0773} & 0.3710 & 0.0796 & 0.3013 \\
RePer-360 (Ours) & ViT-B & 0.0799 & \textbf{0.3577} & \textbf{0.0672} & \textbf{0.2752} \\
\bottomrule 
\end{tabular}
\label{tab:Complex}
\end{table}

\noindent\textbf{Correlation between Model Capacity and Performance Gain.}
RePer-360 remains competitive across backbone scales and outperforms PanDA in most settings, with its advantages becoming more evident for larger backbones. This trend is related to the feature representation capability required by our guidance-based modulation mechanism. Our framework relies on extracting informative geometric cues from the CP branch to guide the ERP representation. Smaller backbones (\textit{e.g.}, ViT-S) inherently possess weaker feature extraction capacity, which limits the quality and richness of the guidance signals derived from cubemap faces. Consequently, the benefit of our dual-branch modulation becomes more pronounced when the backbone is sufficiently powerful (\textit{e.g.}, ViT-B or ViT-L) to capture fine-grained spatial structures present in the cubemap projection.

\subsection{Efficiency and Runtime Analysis}
\label{sec:complexity}

To evaluate the computational efficiency of our framework, we compare model parameters and inference speed with the baseline PanDA-L~\cite{cao2025panda} and the patch-based method MoGe-2~\cite{wang2025moge} in Table~\ref{tab:complexity}. All methods use the ViT-L backbone, and the reported FPS is measured on a single NVIDIA RTX A4000 GPU.

\begin{table}[h]
\centering
\footnotesize
\setlength{\tabcolsep}{10pt}
\caption{\textbf{Complexity Comparison (ViT-L).} RePer-360 introduces moderate overhead while remaining significantly faster than patch-based approaches such as MoGe-2.}
\begin{tabular}{l|c c}
\toprule
Method & Parameters (M) & FPS \\
\midrule
PanDA-L~\cite{cao2025panda} & 336 & 2.59 \\
MoGe-2~\cite{wang2025moge} & 326 & 0.02 \\
RePer-360 (Ours) & 359 & 1.30 \\
\bottomrule
\end{tabular}
\label{tab:complexity}
\end{table}

\noindent\textbf{Parameter Efficiency.}
As shown in Table~\ref{tab:complexity}, RePer-360 introduces only a modest increase in model size, adding 23M parameters compared with PanDA-L ($336\text{M} \rightarrow 359\text{M}$). This corresponds to an increase of approximately $6.8\%$, indicating that the proposed GAG module and the SCAdaLN-Zero mechanism remain lightweight and do not impose a significant storage overhead.

\noindent\textbf{Inference Speed and Trade-offs.}
RePer-360 runs at 1.30 FPS, slower than the single-branch PanDA-L baseline due to the additional ERP-to-CP projection and dual-branch feature extraction. Nevertheless, it remains practical for direct inference and is approximately $\mathbf{65\times}$ faster than MoGe-2, which relies on computationally intensive patch-wise processing.

\noindent\textbf{Comparison with ST$^2$360D~\cite{cao2025st}.}
ST$^2$360D also improves zero-shot depth estimation through a multi-view pipeline, but with higher runtime. Under the reported single-A40 setting, its small variant requires 5.9 s, compared with 0.06 s for PanDA-Small, corresponding to an approximately $98\times$ overhead. Since larger variants are generally not faster than smaller ones, this suggests the high computational cost of multi-view inference. As ST$^2$360D is not publicly available and reports only zero-shot results, we cite its reported numbers. Using its VDA-Large variant, ST$^2$360D reports 0.1153 Abs Rel / 0.4284 RMSE on Matterport3D and 0.1005 Abs Rel / 0.2986 RMSE on Stanford2D3D, whereas RePer-360 achieves 0.1033 / 0.4534 and 0.0630 / 0.2849, respectively. RePer-360 performs better on Stanford2D3D and achieves lower Abs Rel on Matterport3D, while ST$^2$360D reports lower RMSE on Matterport3D.

\subsection{ECCLoss vs. EPNL}
\label{sec:loss_comparison}


\noindent\textbf{Mechanism Difference and Supervision Structure.}
EPNL samples image patches following a Gaussian distribution biased toward the equatorial region and enforces local relative depth consistency. While this strategy improves local supervision, the stochastic sampling process lacks a fixed geometric reference frame. In contrast, ECCLoss introduces a structured projection-based supervision mechanism by transforming the panoramic representation into the cubemap domain. Depth is then supervised on the six cubemap faces, each of which follows standard perspective geometry. This design provides geometrically consistent supervision signals and enables the network to learn spatial relationships under reduced spherical distortion.

\begin{table}[h]
\centering
\footnotesize
\setlength{\tabcolsep}{8pt}
\caption{\textbf{Comparison of Loss Strategies.} Structured supervision via ECCLoss improves performance compared with random patch-based supervision.}
\begin{tabular}{l|c c c}
\toprule
Loss Function & Abs Rel $\downarrow$ & RMSE $\downarrow$ & $\delta_1 \uparrow$ \\
\midrule
w/o Consistency Loss & 0.0716 & 0.3249 & 0.9517 \\
w/ EPNL~\cite{cao2025panda} & 0.0731 & 0.3180 & 0.9548 \\
\rowcolor{gray!10} w/ ECCLoss (Ours) & \textbf{0.0691} & \textbf{0.3164} & \textbf{0.9567} \\
\bottomrule
\end{tabular}
\label{tab:loss_comparison}
\end{table}

\noindent\textbf{Quantitative Results.}
As shown in Table~\ref{tab:loss_comparison}, the baseline without a consistency loss achieves an Abs Rel of 0.0716. Introducing EPNL yields mixed results: although it improves RMSE and $\delta_1$, it slightly worsens Abs Rel. In contrast, ECCLoss provides the best results, reducing Abs Rel to 0.0691. These results indicate that structured, geometrically consistent supervision is more effective for panoramic depth estimation than randomly sampled local constraints.

\subsection{CDF Analysis of Pixel-wise Errors}
\label{sec:cdf}

\begin{figure}[t]
\centering
\includegraphics[width=\textwidth]{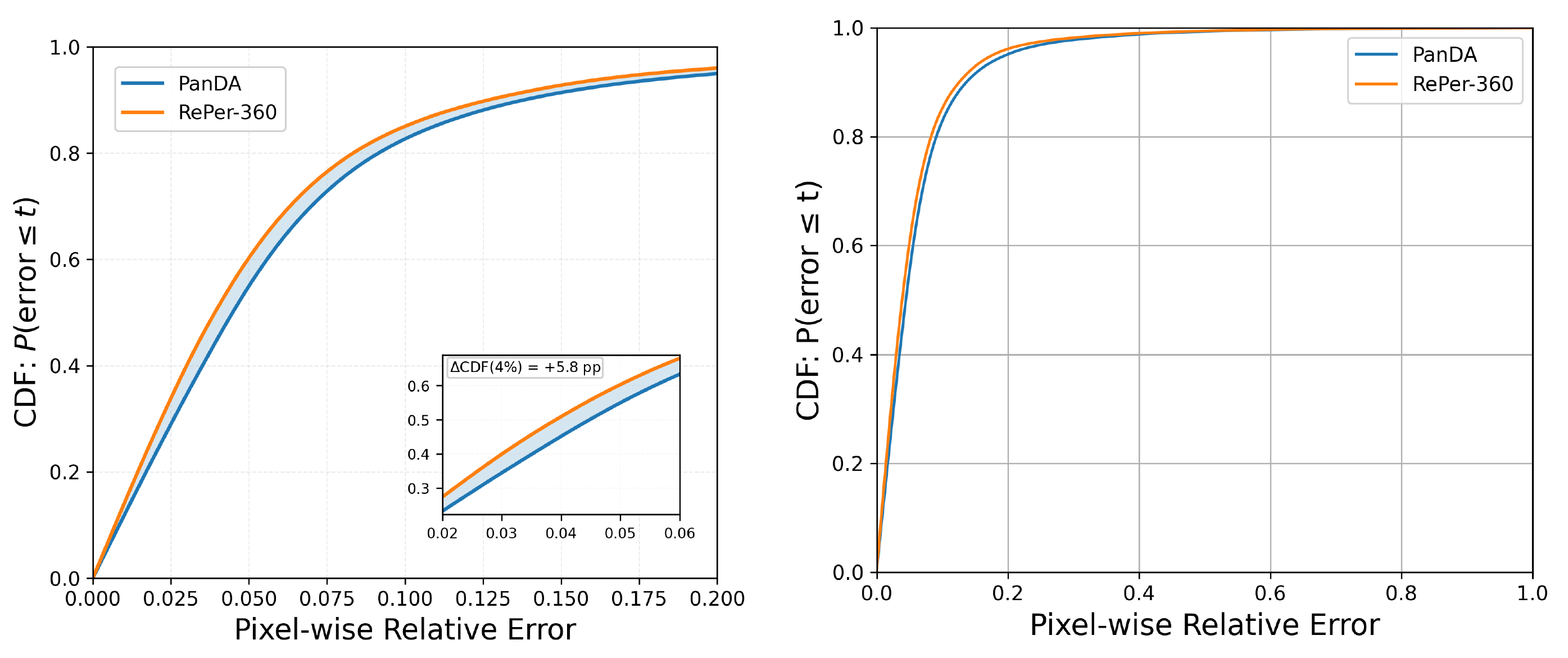}
\caption{\textbf{CDF comparison of pixel-wise relative depth errors.}
RePer-360 consistently achieves higher CDF values than PanDA across error thresholds, indicating that a larger proportion of pixels exhibit lower depth errors. The inset highlights the improvement in the low-error region ($t\approx0.04$), where RePer-360 improves the CDF by approximately $5.8$ percentage points.}
\label{fig:cdf}
\end{figure}

To further analyze the error distribution of different methods, we plot the cumulative distribution function (CDF) of pixel-wise relative depth errors for PanDA and RePer-360, as shown in Fig.~\ref{fig:cdf}. The CDF measures the proportion of pixels whose relative error is below a threshold $t$, providing a distribution-level view of prediction accuracy.

As illustrated in Fig.~\ref{fig:cdf}, the curve of RePer-360 lies consistently above that of PanDA across the evaluated error range. This indicates that a larger fraction of pixels predicted by our method fall within lower error thresholds.

The improvement is particularly evident in the low-error region, which reflects the model’s ability to preserve fine geometric details. For example, at an error threshold of $t=0.04$, RePer-360 achieves an approximately $5.8$ percentage point increase in CDF compared with PanDA.

These results further support that RePer-360 improves global accuracy while reducing pixel-wise depth errors across a broad range of thresholds, leading to more reliable geometric reconstruction.

\subsection{Visualization of the Point Cloud}
\label{sec:vis}

\begin{figure}[t]
\centering
\includegraphics[width=\textwidth]{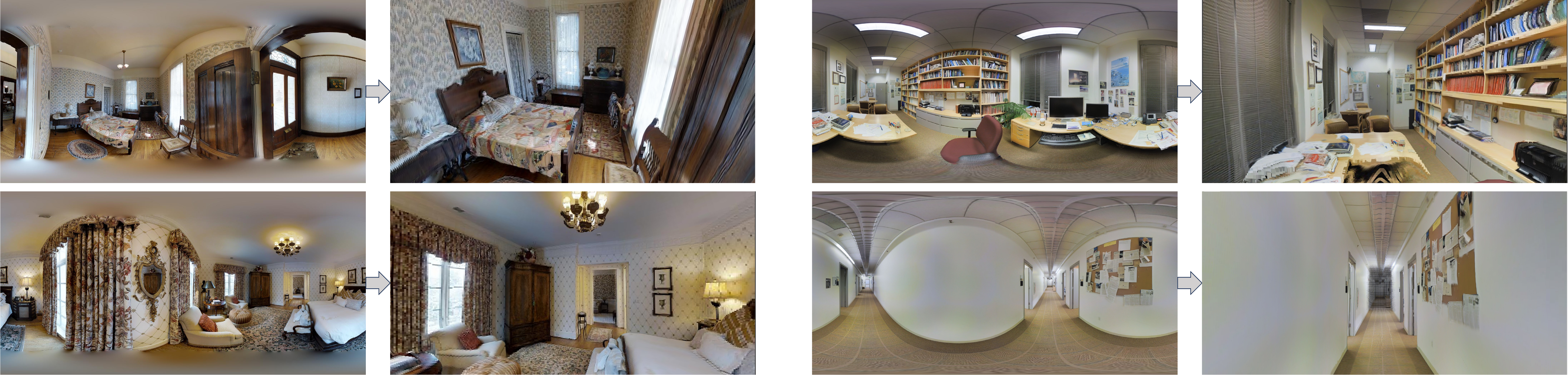}
\caption{Visualization of point clouds generated from depth predictions of RePer-360.}
\label{fig:pointcloud}
\end{figure}

To further illustrate the geometric quality of the predicted depth maps, we project the estimated depth into 3D space to generate omnidirectional point clouds. As shown in Fig.~\ref{fig:pointcloud}, the reconstructed scene geometry demonstrates that RePer-360 preserves coherent spatial structures while maintaining fine architectural details.

\subsection{Qualitative Results on Self-Collected Real-World Panoramas}
\label{sec:self_collected}

We further present qualitative results on self-collected panoramic images captured using a DJI Osmo 360 camera. These samples are collected outside the benchmark datasets and cover diverse real-world indoor and outdoor scenes.

\begin{figure*}[t]
\centering
\includegraphics[width=1\textwidth]{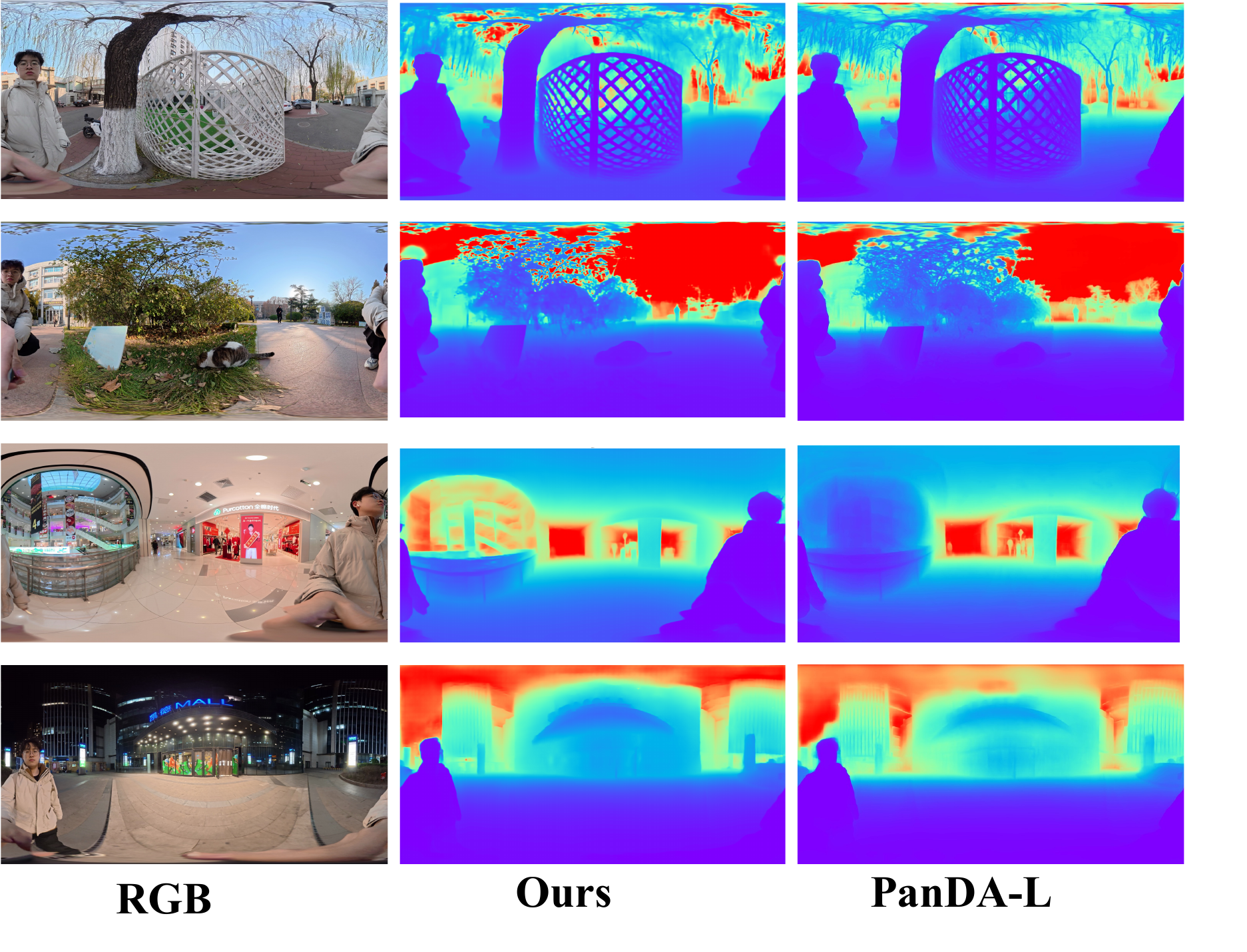}
\caption{\textbf{Qualitative comparison on self-collected panoramas.}
We compare RePer-360 with PanDA-L on panoramas captured using a DJI Osmo 360 camera in diverse real-world scenes. RePer-360 produces coherent depth predictions and preserves clearer boundaries, especially in challenging nighttime and high-distortion regions.}
\label{fig:self_collected}
\end{figure*}

As shown in Fig.~\ref{fig:self_collected}, RePer-360 produces more coherent depth predictions than PanDA-L across diverse scenes. It better preserves structural layout and clearer boundaries in both indoor and outdoor environments, including challenging nighttime cases.
Although these self-collected samples do not provide ground-truth depth, they further suggest that RePer-360 generalizes well to real-world panoramic imagery beyond the benchmark datasets.

\section{Limitation and Future Work}
\label{sec:future}

Despite the additional computational overhead introduced by the dual-branch architecture, which leads to reduced inference speed, and the diminished performance gains observed with smaller backbones, RePer-360 demonstrates the potential of a guidance-based domain adaptation framework. Future work will focus on improving architectural efficiency and extending this mechanism toward a more general framework capable of handling diverse imaging models (\textit{e.g.}, fisheye cameras) and other cross-domain dense prediction tasks.

%
%
\end{document}